\documentclass[11pt]{article}

\usepackage[margin=1in]{geometry}
\usepackage[utf8]{inputenc}
\usepackage[T1]{fontenc}
\usepackage{authblk} % For better author/affiliation handling
\usepackage{microtype}
\usepackage{graphicx}
\usepackage{subcaption}
\usepackage{booktabs}
\usepackage{hyperref}
\hypersetup{
    colorlinks=true,
    linkcolor=blue,
    filecolor=magenta,      
    urlcolor=cyan,
    citecolor=blue,
}
\usepackage{float}
\usepackage{amsmath, amssymb, mathtools, amsthm}
\usepackage{algorithm}
\usepackage{algpseudocode} % 替换了 algorithmic 以支持 \Function 和 \Call
\usepackage[capitalize,noabbrev]{cleveref}

\theoremstyle{plain}

\theoremstyle{definition}

\theoremstyle{remark}

\usepackage[textsize=tiny]{todonotes}

\title{\textbf{Hierarchical Zero-Order Optimization for Deep Neural Networks}}

\author[1]{Sansheng Cao}
\author[2]{Zhengyu Ma\thanks{Corresponding author: mazhy@pcl.ac.cn}}
\author[3]{Yonghong Tian}

\affil[1]{College of Physics, Peking University}
\affil[2]{Pengcheng Laboratory}
\affil[3]{School of Computer Science, Peking University}

\date{}

\begin{document}

\maketitle

\begin{abstract}
Zeroth-order (ZO) optimization has long been favored for its biological plausibility and its capacity to handle non-differentiable objectives, yet its computational complexity has historically limited its application in deep neural networks. Challenging the conventional paradigm that gradients propagate layer-by-layer, we propose Hierarchical Zeroth-Order (HZO) optimization, a novel divide-and-conquer strategy that decomposes the depth dimension of the network. We prove that HZO reduces the query complexity from $O(ML^2)$ to $O(ML \log L)$ for a network of width $M$ and depth $L$, representing a significant leap over existing ZO methodologies. Furthermore, we provide a detailed error analysis showing that HZO maintains numerical stability by operating near the unitary limit ($L_{lip} \approx 1$). Extensive evaluations on CIFAR-10 and ImageNet demonstrate that HZO achieves competitive accuracy compared to backpropagation.
\end{abstract}

\section{Introduction}

The success of deep learning optimization methods relies heavily on the backpropagation (BP) algorithm \cite{1986hintonbp}. However, BP has long faced significant challenges in direct application to biophysical models (e.g., the Hodgkin-Huxley model \cite{hhmodel}) and physical simulation hardware (e.g., optical neural networks \cite{guang1,guang2, guang3}) due to its stringent requirements for differentiability and high precision. Furthermore, the necessity for symmetric weights and differentiability in BP has led to long-standing skepticism regarding its biological plausibility \cite{bpnotreasonable1, bpnotreasonable2}.

As a promising alternative to address these issues, zeroth-order (ZO) optimization has been extensively studied in recent years. ZO optimization has demonstrated remarkable capabilities on low-memory, low-precision, and analog devices. Specifically, the memory footprint of ZO methods is often half or even less than that of BP, making it feasible to fine-tune large-scale models on edge devices \cite{2024llmfinetuning}. Additionally, ZO methods based on quantized networks enable fine-tuning without the need for high-precision BP signals \cite{zolianghua}, while significant progress has also been made in training optical neural networks using ZO approaches \cite{zoguang1,zoguang2}.

However, the core challenge of ZO methods—the "curse of dimensionality"—has long hindered their broader adoption. For a network with depth $L$ and average width $M$, the time complexity of traditional ZO methods is $O(M^2L^2)$, which is computationally prohibitive for deep architectures with thousands of layers and neurons. To mitigate this, several strategies have been introduced. Notably, \cite{ren2022scaling} leveraged biological local learning rules to propose activation-space perturbations with block-wise losses. By shifting the perturbation target from weights to neurons, the time complexity was reduced from $O(M^2L^2)$ to $O(ML^2)$, with block-wise strategies offering a further constant-factor reduction. Furthermore, low-rank perturbations have achieved additional efficiency gains by restricting perturbations to a low-rank subspace \cite{yu2024subzero}, and DeepZero has halved the forward-pass complexity through feature reuse \cite{chen2023deepzero}Nevertheless, for ultra-deep networks, these methods still fail to resolve the fundamental $O(L^2)$ complexity bottleneck.

To this end, this paper proposes a novel Hierarchical Zeroth-Order (HZO) optimization method. Inspired by the hierarchical structure of the biological brain \cite{frank2023hierarchicalinbiology, theves2021learning, liu2022electrophysiological} and the concept of renormalization in physics \cite{beny2013deep, mehta2014exact}, we introduce a divide-and-conquer strategy along the depth dimension. This approach successfully reduces the time complexity of ZO optimization to $O(ML \log L)$. Furthermore, we provide a detailed analysis of the error propagation and memory overhead of HZO, comparing it against traditional ZO, activation-perturbation methods, and BP to demonstrate the superior performance and robustness of the HZO framework.

\section{Related Work}

\subsection{ZO in Fine-tuning}

Zeroth-order (ZO) optimization has recently regained attention as a practical alternative to backpropagation in scenarios where gradients are unavailable, unreliable, or prohibitively expensive to compute. A particularly active line of work applies ZO methods to large-scale model fine-tuning under strict hardware constraints, such as limited GPU memory, low numerical precision, or restricted access to automatic differentiation.

Representative examples include MeZO~\cite{malladi2024finetuninglanguagemodelsjust} and the work of~\cite{mi2025fastllmfinetuningzerothorder}, which demonstrate that ZO-based updates can effectively adapt large pretrained language models using only forward passes and a small memory footprint. These approaches highlight the practical viability of ZO optimization in deployment-oriented settings, especially when fine-tuning foundation models on edge devices or consumer-grade hardware.

Despite their empirical success, these methods incur a significantly higher query complexity compared to gradient-based fine-tuning, as each parameter update requires multiple function evaluations. As a result, they are primarily limited to fine-tuning regimes with a small number of optimization steps and are generally considered impractical for training deep networks from scratch.

\subsection{ZO Computation Complexity}

To address the prohibitive cost of naive ZO optimization, a substantial body of prior work has focused on reducing the dimensionality of perturbations, predominantly along the width dimension of neural networks. For instance, \cite{ren2022scaling} proposed neuron-wise perturbations, which replace full-parameter perturbations with structured updates at the neuron level, dramatically reducing the dependence on network width.

In parallel, advanced sampling strategies and optimizer designs have been introduced to further improve efficiency. Adaptive ZO optimizers~\cite{cao2024acceleratedgradientmethodconvex} adjust perturbation magnitudes or directions based on historical information, while entropy-driven sampling approaches~\cite{koslik2024hiddensemimarkovmodelsinhomogeneous} aim to concentrate queries in more informative subspaces. These methods effectively reduce the constant factors in ZO complexity and improve empirical convergence.

Building on these advances, DeepZero~\cite{chen2023deepzero} demonstrated for the first time that zeroth-order methods can train ImageNet-scale models from scratch, substantially narrowing the performance gap between ZO optimization and backpropagation. Nevertheless, despite these impressive results, existing ZO methods still suffer from unfavorable scaling with respect to network depth. Since they primarily reduce complexity in the width dimension—through weights, neurons, or low-dimensional subspaces~\cite{yu2024subzero}—their query complexity typically grows quadratically with depth.

In contrast, HZO is orthogonal to these approaches. Rather than compressing perturbations within layers, HZO directly decomposes the depth dimension itself via a hierarchical divide-and-conquer strategy. This enables a reduction of query complexity to $O(ML\log L)$, fundamentally altering the depth dependence of zeroth-order optimization.

\subsection{ZO and Biologically Plausible Algorithms}

From the perspective of biological plausibility, several alternative learning paradigms have been proposed to address long-standing criticisms of backpropagation, most notably the requirement of exact weight symmetry between forward and backward passes. Feedback Alignment~\cite{lillicrap2014randomfeedbackweightssupport} replaces transposed weights with fixed random feedback connections, while Target Propagation~\cite{bengio2014autoencodersprovidecreditassignment} assigns credit through layer-wise reconstruction targets instead of explicit gradient signals.

Although conceptually appealing and more biologically grounded, these methods typically struggle to scale to deep architectures. A common failure mode is the accumulation of approximation errors across layers, which leads to unstable training dynamics and degraded performance as depth increases.

HZO bridges this gap by combining the biological appeal of zeroth-order learning with a principled mechanism for preserving global loss coherence across layers. By hierarchically coordinating local perturbations, HZO mitigates error accumulation along depth and achieves performance comparable to backpropagation on standard deep learning benchmarks.

\section{Method}

\subsection{Problem Description}

We consider a deep neural network with $L$ layers. Use $F(\mathbf{x}) = f_L \circ f_{L-1} \circ \dots \circ f_1(\mathbf{x})$ to formalize the network. For layer $l$, the activation function is given by $\mathbf{a}_l = \sigma(\mathbf{W}_l \mathbf{a}_{l-1} + \mathbf{b}_l)$, where $\mathbf{W}_l \in \mathbb{R}^{M_{l-1} \times M_l}$ represents the weight matrix. We use $M=\sup{M_l}$ to simplify subsequent discussion. The objective is to minimize the empirical risk $\mathcal{L}(F(\mathbf{x}), \mathbf{y})$ over a dataset $\{(\mathbf{x}_i, \mathbf{y}_i)\}_{i=1}^N$. 

In our problem, we no longer demand that $\sigma$ is explicitly differentiable as normal BP method did. Differentiablility will only be introduced in the next chapter when discussing errors. 

\subsection{Hierarchical Zeroth-Order Optimizer (HZO)}

\subsubsection{Formal Definition of Subnetworks}
To establish a recursive optimization framework, we first define the architecture of a deep neural network as a composition of layers. Let a network of depth $L$ be represented as $F(\mathbf{x}) = f_L \circ f_{L-1} \circ \dots \circ f_1(\mathbf{x})$. We formally define a subnetwork $\mathcal{N}_{i:j}$ spanning from layer $i$ to layer $j$ ($1 \le i \le j \le L$) as:
\begin{equation}
    \mathcal{N}_{i:j}(\mathbf{a}_{i-1}) = f_j(f_{j-1}(\dots f_i(\mathbf{a}_{i-1})\dots))
\end{equation}
where $\mathbf{a}_{i-1}$ is the input activation from the preceding layer. The forward pass through this subnetwork generates the activation $\mathbf{a}_j = \mathcal{N}_{i:j}(\mathbf{a}_{i-1})$. We define the target signal $\mathbf{T}_j$ at layer $j$ as the derivative of the global loss function $\mathcal{L}$ with respect to the activation $\mathbf{a}_j$:
\begin{equation}
    \mathbf{T}_j = \frac{\partial \mathcal{L}}{\partial \mathbf{a}_j}
\end{equation}

\subsubsection{Recursive Bisection and Target Propagation}
The HZO strategy employs a divide-and-conquer approach along the depth dimension. For any subnetwork $\mathcal{N}_{i:j}$ with $j > i$, we perform a bisection at the midpoint $k = \lfloor (i+j)/2 \rfloor$, splitting the structure into $\mathcal{N}_{left} = \mathcal{N}_{i:k}$ and $\mathcal{N}_{right} = \mathcal{N}_{k+1:j}$. 

To propagate the target signal $\mathbf{T}_j$ back to the intermediate layer $k$, we estimate the Jacobian matrix $\mathbf{J} = \frac{\partial \mathbf{a}_j}{\partial \mathbf{a}_k}$ of the right subnetwork $\mathcal{N}_{right}$ using the bidirectional difference method:
\begin{equation}
    \mathbf{J}_{mn} \approx \frac{[\mathcal{N}_{right}(\mathbf{a}_k + \epsilon \mathbf{e}_n)]_m - [\mathcal{N}_{right}(\mathbf{a}_k - \epsilon \mathbf{e}_n)]_m}{2\epsilon}
\end{equation}
where $\epsilon$ is a small perturbation constant and $\mathbf{e}_n$ is the $n$-th basis vector. The target signal for the intermediate layer is then computed as:
\begin{equation}
    \mathbf{T}_k = \mathbf{J}^\top \mathbf{T}_j
\end{equation}
This procedure is applied recursively until the depth of the subnetwork reaches unity ($i=j$).

\subsubsection{Local Learning Rule and Gradient Equivalence}
When the recursion reaches a single layer $\mathcal{N}_{i:i}$, we apply the Delta Rule\cite{hanson1990stochastic} to update the local weights $\mathbf{W}_i$ . The weight update $\Delta \mathbf{W}_i$ is given by the outer product of the input activation and the target signal:
\begin{equation}
    \Delta \mathbf{W}_i = -\eta \mathbf{T}_i \mathbf{a}_{i-1}^\top
\end{equation}
where $\eta$ is the learning rate. Technically, we use local auto-differentiation in Pytorch instead due to its exactly equivalence.

\subsection{Algorithm}
The pseudo-code of the HZO algorithm is as follows:
\begin{algorithm}[H]
\caption{Hierarchical Zeroth-Order Optimization}
\label{alg:hzo}
\begin{algorithmic}[1]
\Require Network $\mathcal{N}_{1:L}$, Input $\mathbf{x}$, Target $\mathbf{T}_L = \frac{\partial \mathcal{L}}{\partial \mathbf{a}_L}$, Perturbation $\epsilon$, Learning rate $\eta$
\Ensure Updated weights $\mathbf{W}_{1:L}$

\Function{HZO}{$\mathcal{N}_{i:j}, \mathbf{a}_{i-1}, \mathbf{T}_j$}
    \If{$i = j$} \label{alg:base_case_start}
        \State Compute activation $\mathbf{a}_i = f_i(\mathbf{a}_{i-1})$
        \State Update weights: $\mathbf{W}_i \leftarrow \mathbf{W}_i - \eta (\mathbf{T}_i \mathbf{a}_{i-1}^\top)$
    \Else
        \State $k \gets \lfloor (i+j)/2 \rfloor$
        \State $\mathcal{N}_{left} \gets \mathcal{N}_{i:k}$, $\mathcal{N}_{right} \gets \mathcal{N}_{k+1:j}$
        \State $\mathbf{a}_k \gets \mathcal{N}_{left}(\mathbf{a}_{i-1})$
        \State
        \For{each neuron $m$ in layer $k$}
            \State $\mathbf{a}_k^+ \gets \mathbf{a}_k + \epsilon \mathbf{e}_m, \quad \mathbf{a}_k^- \gets \mathbf{a}_k - \epsilon \mathbf{e}_m$
            \State $\mathbf{J}_m \gets \frac{\mathcal{N}_{right}(\mathbf{a}_k^+) - \mathcal{N}_{right}(\mathbf{a}_k^-)}{2\epsilon}$
        \EndFor
        \State $\mathbf{T}_k \gets \mathbf{J}^\top \mathbf{T}_j$
        \State
        \State \Call{HZO}{$\mathcal{N}_{right}, \mathbf{a}_k, \mathbf{T}_j$}
        \State \Call{HZO}{$\mathcal{N}_{left}, \mathbf{a}_{i-1}, \mathbf{T}_k$}
    \EndIf
\EndFunction
\State
\State \Call{HZO}{$\mathcal{N}_{1:L}, \mathbf{x}, \mathbf{T}_L$}
\end{algorithmic}
\end{algorithm}

\subsection{Spatial Parallel Perturbation for Convolutional Layers}

To extend the HZO framework to high-dimensional Convolutional Neural Networks (CNNs), we introduce the Spatial Parallel Perturbation (SPP) strategy. In a standard zeroth-order approach, estimating the gradient of a feature map $\mathbf{A} \in \mathbb{R}^{C \times H \times W}$ requires $\mathcal{O}(H \times W)$ forward queries, which is computationally prohibitive for large-scale images. 

Exploiting the local connectivity of convolutional kernels, we observe that perturbations at two spatial locations $(h, w)$ and $(h', w')$ are independent if their respective receptive fields do not overlap in the subsequent layer. Formally, let $R$ be the effective receptive field size of the current block. We define a set of spatially disjoint indices $\mathcal{S}$ such that:
\begin{equation}
\forall (h, w), (h', w') \in \mathcal{S}, \quad \|(h, w) - (h', w')\|_\infty \geq R
\end{equation}

By utilizing SPP, the query complexity per layer is reduced from $\mathcal{O}(H \times W)$ to a constant $\mathcal{O}(R^2)$, effectively decoupling the computational cost from the input resolution. This optimization ensures that HZO remains efficient even when processing high-resolution inputs in ImageNet-scale tasks.

\section{Theoretical Analysis}
In this chapter, we conducted a rigorous mathematical analysis of the HZO method and proved its equivalence to the BP method, its superiority in terms of complexity, and its controllability of errors.

\subsection{Equivalence to Backpropagation}
\paragraph{Theorem 1 (Gradient Equivalence)} The recursive application of the Jacobian-target product in HZO is equivalent to the analytical gradient computation in Backpropagation (BP).

\textit{Proof Sketch.} In BP, the gradient with respect to layer $k$ is given by:
\begin{equation*}
    \frac{\partial \mathcal{L}}{\partial \mathbf{a}_k} = \left(\frac{\partial \mathbf{a}_j}{\partial \mathbf{a}_k}\right)^\top \frac{\partial \mathcal{L}}{\partial \mathbf{a}_j}
\end{equation*}
By defining $\mathbf{J}$ as the Jacobian $\frac{\partial \mathbf{a}_j}{\partial \mathbf{a}_k}$ and $\mathbf{T}_j$ as the target gradient $\frac{\partial \mathcal{L}}{\partial \mathbf{a}_j}$, the operation:
\begin{equation*}
    \mathbf{T}_k = \mathbf{J}^\top \mathbf{T}_j
\end{equation*}
directly recovers the standard chain rule. 

Furthermore, considering the layer activation $\mathbf{a}_i = \sigma(\mathbf{W}_i \mathbf{a}_{i-1})$, the gradient with respect to the weights is:
\begin{equation*}
    \frac{\partial \mathcal{L}}{\partial \mathbf{W}_i} = \frac{\partial \mathcal{L}}{\partial \mathbf{a}_i} \frac{\partial \mathbf{a}_i}{\partial \mathbf{W}_i}
\end{equation*}
For standard linear operations, $\frac{\partial \mathbf{a}_i}{\partial \mathbf{W}_i}$ yields $\mathbf{a}_{i-1}^\top$. Consequently, the HZO-derived update:
\begin{equation*}
    \Delta \mathbf{W}_i \propto \mathbf{T}_i \mathbf{a}_{i-1}^\top
\end{equation*}
exactly matches the direction of the BP update, effectively implementing the Delta Rule without explicit global backward propagation.

\subsection{Complexity Analysis and Proof}

To rigorously establish the computational efficiency of the Hierarchical Zeroth-Order (HZO) optimizer, we analyze the total number of forward passes required during one complete optimization cycle. 

\subsubsection{Notations and Preliminaries}
Let $L$ denote the total number of layers in the network and $M$ denote the average width (number of neurons) of each layer. We assume that a single forward pass through a sub-network of depth $l$ has a computational cost proportional to $l \cdot O(M^2)$, where $O(M^2)$ represents the overhead of a matrix-vector multiplication at each layer. 

\subsubsection{Recurrence Relation for HZO}
The HZO algorithm optimizes the network via a recursive bisection strategy. For a sub-network of depth $L$, the total computational cost $T(L)$ is composed of the costs of optimizing two bisected sub-networks of depth $L/2$ and the cost of estimating the Jacobian at the bisection point. The recurrence relation is defined as:
\begin{equation}
    T(L) = 2 \cdot T(L/2) + C_{Jacobian}(L/2)
\end{equation}
where $C_{Jacobian}(L/2)$ represents the cost of estimating the Jacobian of the right sub-network (depth $L/2$) with respect to the $M$ neurons at the bisection point.

\subsubsection{Level-wise Summation and Proof}
To solve this recurrence, we analyze the computational workload at each level of the recursion tree. Let $k$ denote the level of recursion, where $k = 0, 1, \dots, \log_2 L - 1$.

\begin{enumerate}
    \item \textbf{Number of Sub-problems}: At level $k$, there are $2^k$ independent sub-networks being optimized.
    \item \textbf{Sub-network Depth}: Each sub-problem at level $k$ involves a sub-network of effective depth $L/2^k$.
    \item \textbf{Jacobian Estimation Cost}: For each sub-problem, we estimate the Jacobian of the right partition (depth $L/2^{k+1}$) by perturbing $M$ neurons. Using the bidirectional difference method, this requires $2M$ forward passes of length $L/2^{k+1}$, plus one reference forward pass. The cost per sub-problem at level $k$ is:
    \begin{equation}
        \text{Cost}_{sub}(k) \approx 2M \cdot \left( \frac{L}{2^{k+1}} \right) = \frac{ML}{2^k}
    \end{equation}
\end{enumerate}

The total cost at level $k$, denoted by $W_k$, is the product of the number of sub-problems and the cost per sub-problem:
\begin{equation}
    W_k = 2^k \cdot \left( \frac{ML}{2^k} \right) = ML
\end{equation}

\textbf{Total Complexity Calculation}: Summing the costs across all $\log_2 L$ levels of recursion, we obtain the total complexity $T(L)$ :
\begin{equation}
    T(L) = \sum_{k=0}^{\log_2 L - 1} W_k = \sum_{k=0}^{\log_2 L - 1} ML  = ML \log_2 L
\end{equation}

\textbf{Theorem 2 (Time Complexity):} The query complexity of the HZO optimizer is $O(ML \log L)$ forward passes.

\subsection{Error Accumulation Analysis}

In this section, we analyze the numerical stability of the HZO optimizer. We show that while HZO achieves significant computational speedup, its gradient estimation error follows the same exponential scaling laws as traditional Zeroth-Order (ZO) methods, necessitating specific architectural constraints for ultra-deep networks.

\subsubsection{The Geometry of Error Propagation}
Consider a deep neural network of depth $L$, represented as the composite function $F(\mathbf{x}) = f_L \circ f_{L-1} \circ \dots \circ f_1(\mathbf{x})$. The accuracy of any zeroth-order gradient estimate depends on the second-order Taylor expansion of this composite function. Based on the operator norm of the Hessian $\frac{\partial^2 F}{\partial \mathbf{x}^2}$, we derive the second-order term as a summation of layer-wise contributions:
\begin{equation}
    \frac{\partial^2 F}{\partial \mathbf{x}^2} = \sum_{k=1}^L \left[ \left( \prod_{i=k+1}^L J_i \right) \cdot H_k \cdot \left( \prod_{j=1}^{k-1} J_j \right)^2 \right]
\end{equation}
where $J_k = \frac{\partial f_k}{\partial f_{k-1}}$ is the Jacobian and $H_k = \frac{\partial^2 f_k}{\partial f_{k-1}^2}$ is the Hessian of the $k$-th layer. Assuming each layer $f_k$ is $L_{lip}$-Lipschitz continuous and $\beta$-smooth, the spectral norm of the estimation error $\mathcal{E}$ for a perturbation $\delta$ is bounded by:
\begin{equation}
    \|\mathcal{E}\| \le O\left( \beta \cdot L_{lip}^{2L} \cdot \|\delta\|^2 \right)
\end{equation}
This formulation reveals a fundamental property: the estimation error of zeroth-order methods inherently grows exponentially with the network depth $L$.

\subsubsection{Error Consistency and Stability}
A critical finding of our work is that the recursive bisection strategy of HZO does not introduce additional error magnitudes compared to global perturbation methods.

\textbf{Theorem 3 (Error Magnitude Consistency).} \textit{The gradient estimation error of the HZO optimizer is in the same order of magnitude as that of direct Zeroth-Order methods.}

\textit{Proof Sketch.} In HZO, the final gradient is recovered through $\log_2 L$ levels of recursive Jacobian-target products. Although each recursive step involves a finite-difference approximation, the total dependency path of the composite function remains $L$. Consequently, the cumulative error exhibits the same $O(L_{lip}^{2L})$ scaling as a direct perturbation of the entire network. The detailed algebraic derivation of this bound is provided in Appendix A.

\subsubsection{The Unitary Limit: $L_{lip} \approx 1$}
The exponential dependence on $L_{lip}$ implies that for ultra-deep architectures, zeroth-order optimization is only stable when the network operates near the unitary limit, where $L_{lip} \approx 1$. If $L_{lip} > 1$, the estimation error explodes, overwhelming the first-order gradient signal. 

This theoretical insight explains why HZO performs robustly in architectures that maintain $L_{lip} \approx 1$, such as those utilizing residual connections or orthogonal initialization. Under such constraints, the error remains $O(\beta \|\delta\|^2)$, which is depth-invariant and ensures stable convergence for networks with thousands of layers.

\section{Experiment}
\subsection{Evaluation via Cosine Similarity}
In this section, we evaluate the gradient estimation accuracy of the HZO method by comparing it with the gradients computed via Backpropagation (BP). We analyze the cosine similarity between these two gradient vectors across various network depths ($L \in \{16, 32, 64\}$) and architectures (Plain CNN and ResNet). Cosine similarity is employed to measure the directional alignment between two vectors in a high-dimensional space. The value ranges from $-1$ to $1$, where a value approaching $1$ indicates that the two vectors point in nearly identical directions, signifying a highly accurate gradient estimation. The cosine similarity $\rho$ is defined as follows:
\begin{equation}\rho = \frac{\mathbf{g}_{BP} \cdot \hat{\mathbf{g}}_{HZO}}{|\mathbf{g}_{BP}| |\hat{\mathbf{g}}_{HZO}|}\end{equation}
where $\mathbf{g}_{BP}$ represents the ground-truth gradient vector calculated by BP, and $\hat{\mathbf{g}}_{HZO}$ denotes the gradient vector estimated by the HZO method for the same layer. By examining the evolution of $\rho$ as depth $L$ increases, we can empirically determine the stability and scalability of the hierarchical zeroth-order estimation

\begin{figure}[H]
    \centering
    \includegraphics[width=0.6\linewidth]{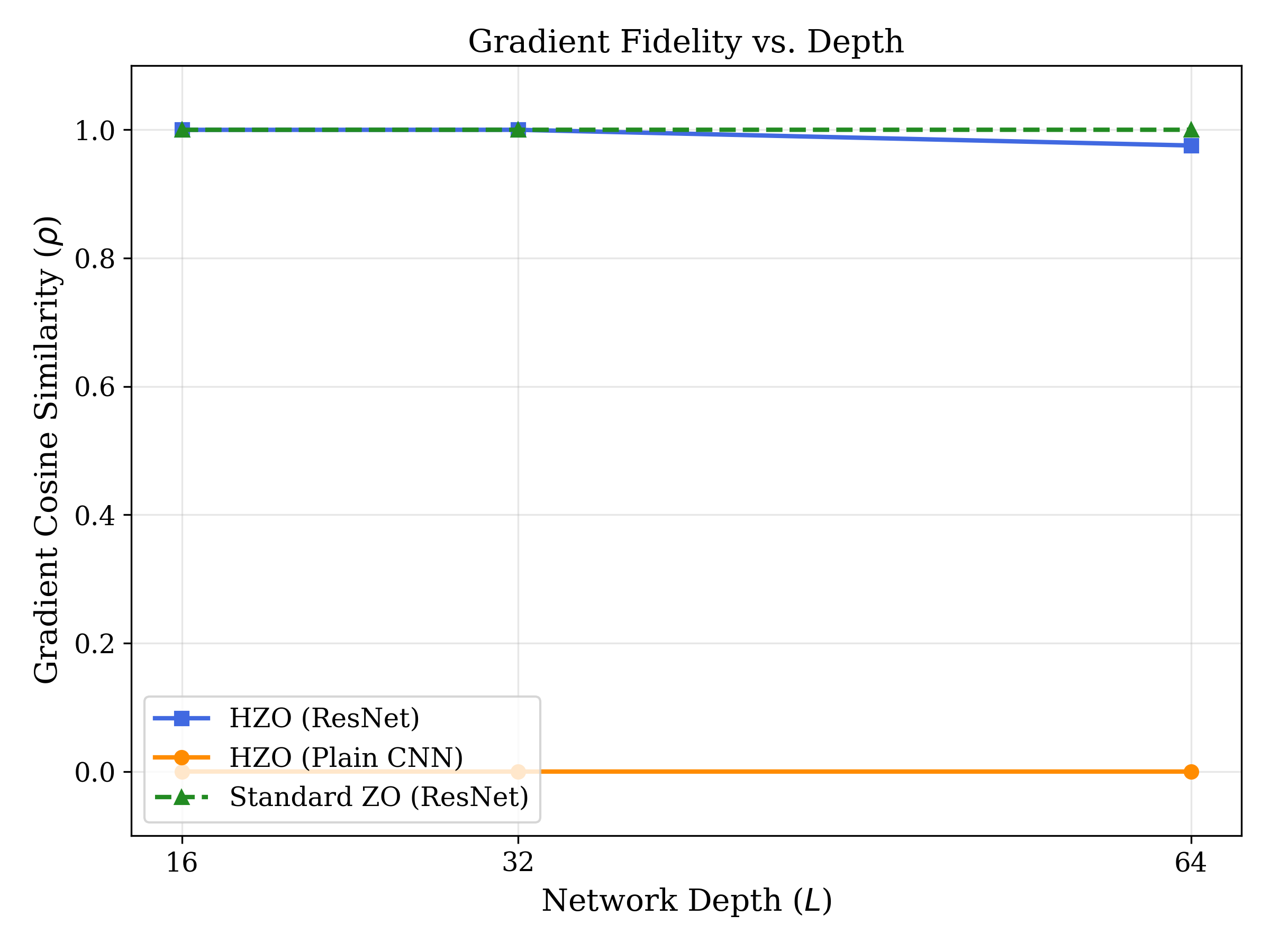}
    \caption{\textbf{Gradient Cosine Similarity vs. Network Depth.} We compare the cosine similarity of gradients estimated by HZO and Standard ZO across various depths ($L \in \{16, 32, 64\}$). HZO combined with ResNet maintains near-perfect directional alignment ($\rho \approx 1.0$) even at $L=64$, whereas the cosine similarity of Plain CNNs remains low.}
    \label{fig:cos_sim_depth}
\end{figure}

\subsection{Performance on CIFAR-10 and ImageNet}
To evaluate the effectiveness of HZO, we first conduct experiments on the CIFAR-10 dataset. For a fair comparison, we adopt the same network architecture as DeepZero\cite{chen2023deepzero}.The training process is optimized using the Adam optimizer to ensure efficient convergence. To maintain high computational throughput and consistent timing benchmarks, all experiments are executed on a high-performance computing node equipped with an NVIDIA A100 GPU. 

Our method achieves a peak test accuracy of 74.2\% after 8 hours of training on CIFAR10 dataset \cite{cifar10}, outperforming all existing non-backpropagation (non-BP) paradigms to the best of our knowledge. (We directly use the result in \cite{chen2023deepzero}, Although we are using an A100, compared with their 4 V100s) A critical factor contributing to this performance is the high cosine similarity; the cosine similarity between HZO-estimated gradients and ground-truth gradients remains consistently above 0.95 throughout the entire training process. This remarkably high similarity validates the superior stability of the hierarchical bisection strategy in deep architectures. Detailed comparisons with prior art are summarized in Table \ref{tab:cifar_results}.

\begin{table}[h]
\centering
\caption{Performance and time consumption of HZO and other BP free method}
\label{tab:cifar_results}
\begin{tabular}{l|c|c|c|c}
\hline
\hline
Method & DeepZero & FA & Align-ada & \textbf{HZO} \\
\hline
Accuracy & 64.1 & 46.5  & 49.9 & \textbf{74.2} \\
\hline
Time (h) & 28.15 & 0.36 & 0.48 & 8.14 \\
\hline
\hline
\end{tabular}
\end{table}

\begin{figure}[H]
    \centering
    \includegraphics[width=0.6\linewidth]{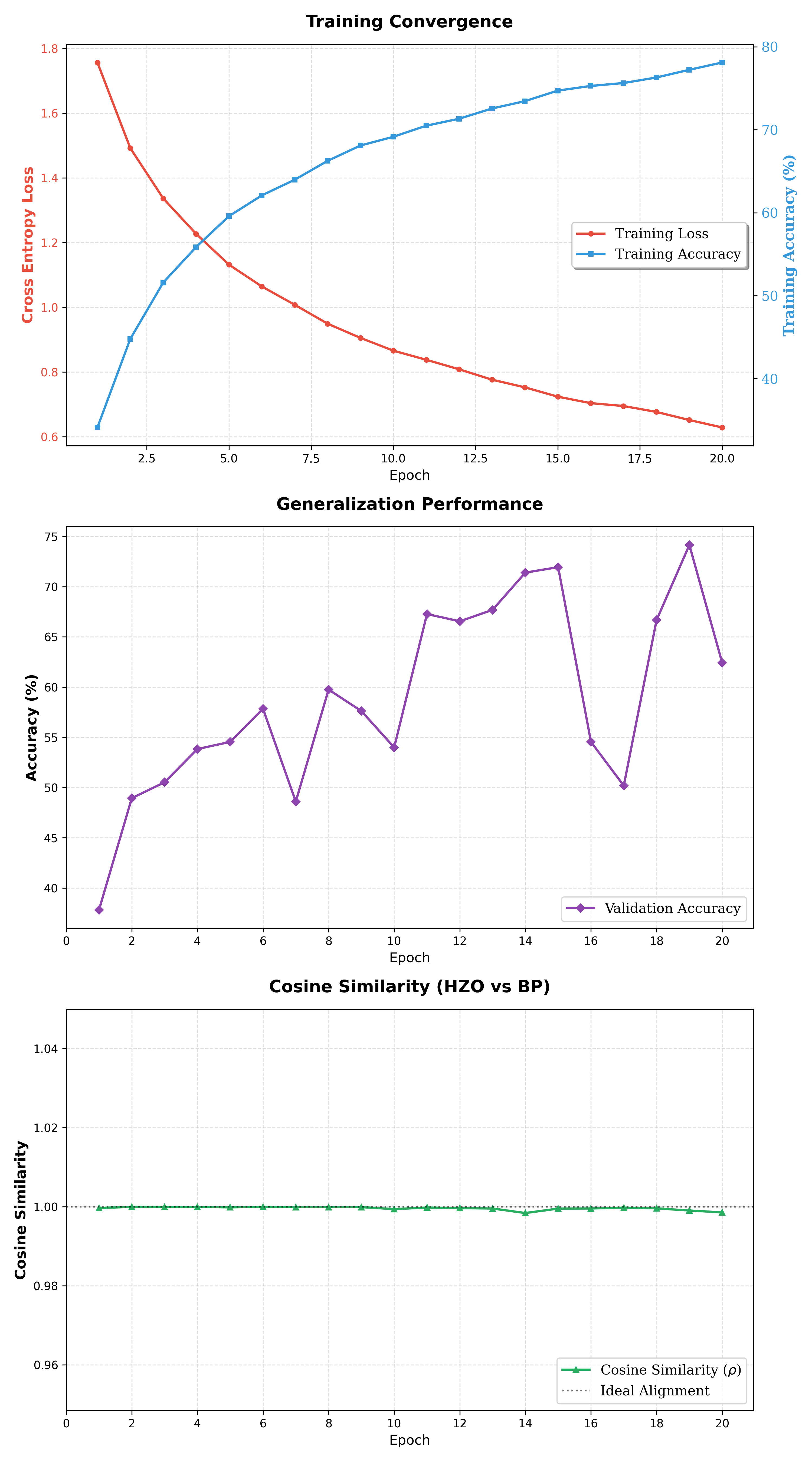}
    \caption{Experimental Results on CIFAR10 Dataset. (Top) The training loss in red color and accuracy in blue color. (Middle) Validation accuracy of HZO method. (Bottom) Cosine similarity between HZO and BP methods.}
    \label{fig:cifar10}
\end{figure}

\begin{figure}[H]
    \centering
    \includegraphics[width=0.6\linewidth]{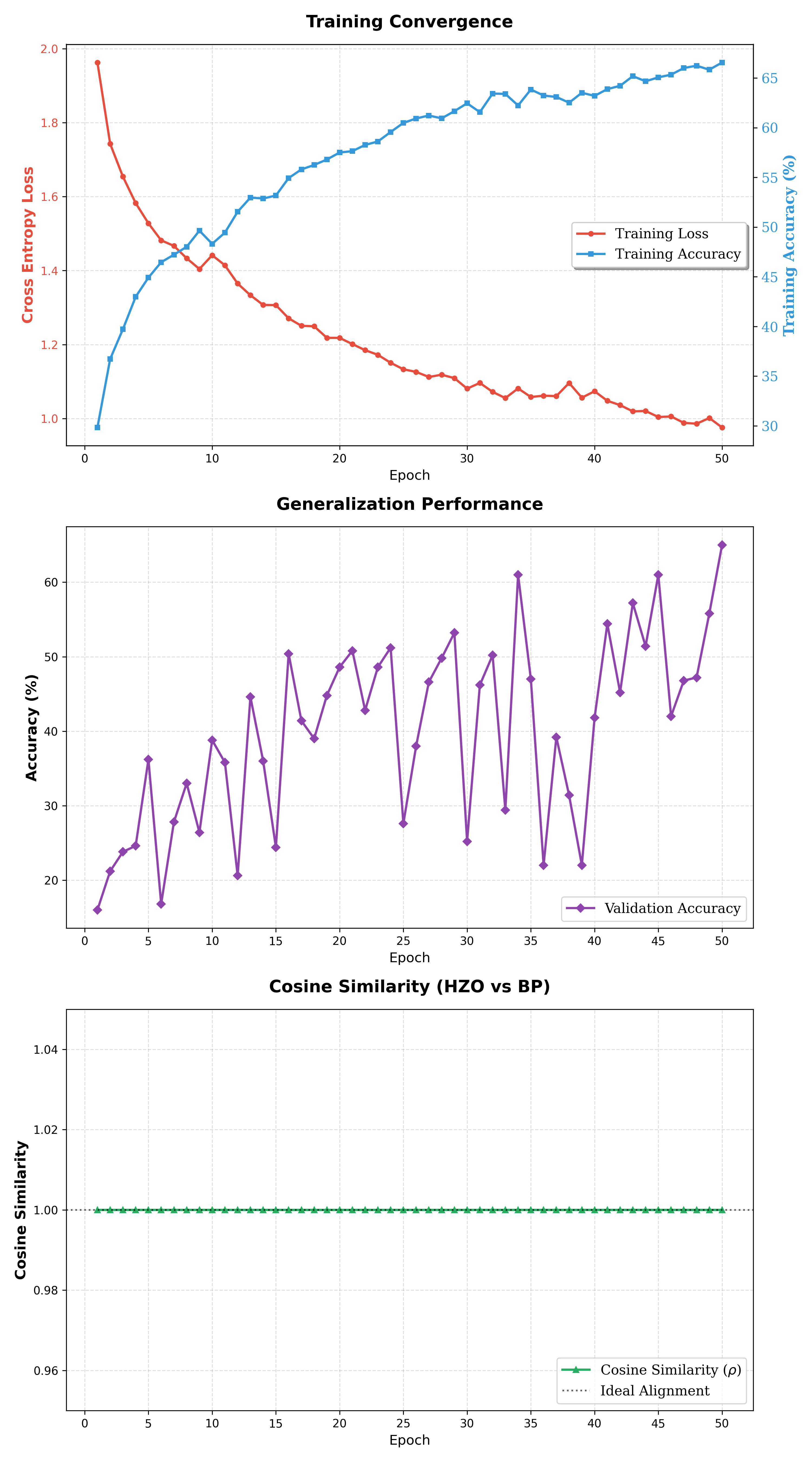}
    \caption{Experimental Results on ImageNet Dataset. (Top) The training loss in red color and accuracy in blue color. (Middle) Validation accuracy of HZO method. (Bottom) Cosine similarity between HZO and BP methods.}
    \label{fig:imagenet}
\end{figure}

To further demonstrate the potential of HZO for large-scale and high-resolution tasks, we extended our evaluation to a 10-class subset of ImageNet \cite{deng2009imagenet} and an accuracy rate of 65.0\% was achieved on the test set. The complete training trajectories for both CIFAR-10 and ImageNet-10 are visualized in \ref{fig:cifar10}, \ref{fig:imagenet}.

As shown in the accuracy curves, while we observed certain fluctuations in the test set, empirical analysis suggests that this instability stems from the internal state dynamics of the Adam optimizer (e.g., the accumulation of second-order moments under zeroth-order noise) rather than a fundamental flaw in the HZO gradient estimation itself. Despite these fluctuations, the results indicate that HZO possesses the scalability required for high-dimensional optimization. Compared to current non-BP methods, HZO offers significant improvements in both convergence speed and final accuracy, marking a substantial step toward practical non-BP training of deep neural networks.

\subsection{Computational Efficiency}

In this section, we evaluate the computational efficiency of HZO by conducting depth-scaling tests on multi-layer perceptron (MLP) architectures. We compare the wall-clock time per gradient estimation of HZO against three representative paradigms: Standard Zeroth-Order (ZO-SGD), Neuron-wise Perturbation (DeepZero), and Backpropagation (BP).

We construct a series of deep MLP networks with hidden layers $L \in \{4, 16, 32, 64\}$, each with a constant hidden dimension. This setup allows us to isolate the impact of network depth on the time complexity of gradient estimation.As demonstrated in Figure \ref{fig:scaling}, the experimental results reveal a significant performance hierarchy:

\begin{figure}[H]
    \centering
    \includegraphics[width=0.7\linewidth]{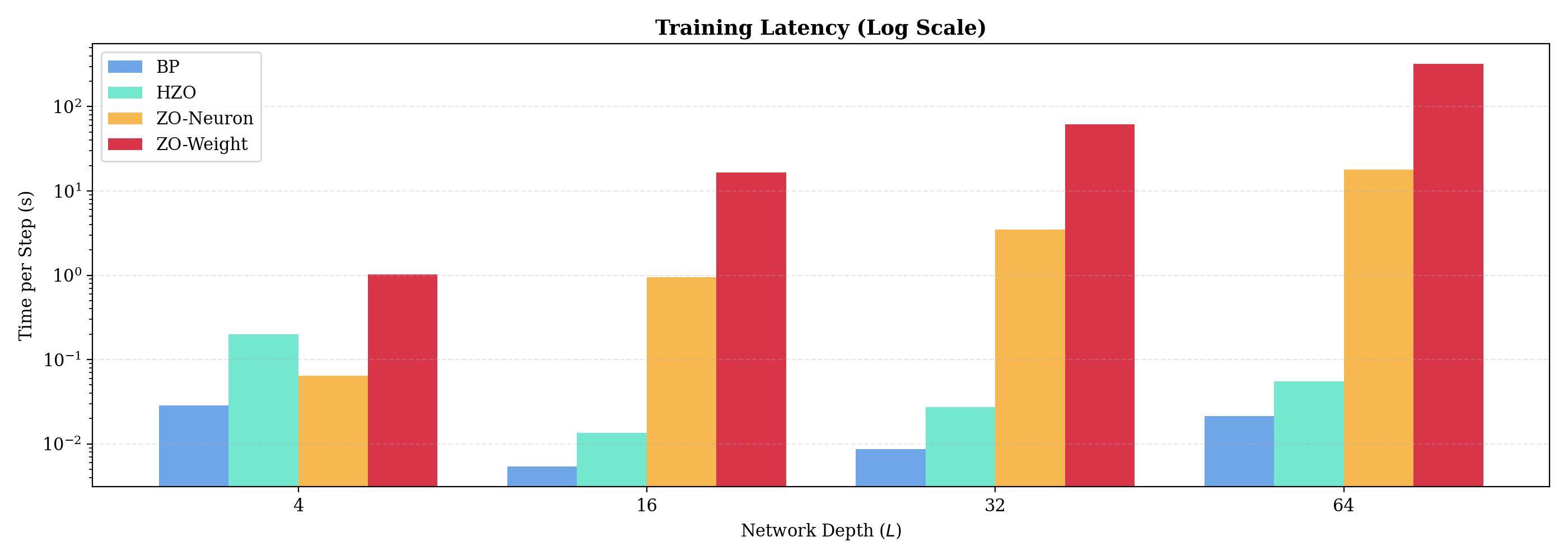}
    \caption{Comparison of computational complexity}
    \label{fig:scaling}
\end{figure}

\subsection{Ablation Study: Numerical Precision and Activation Function}
In this section, we investigate how HZO's gradient cosine similarity is influenced by the choice of activation functions and numerical precision. Our theoretical analysis in Section 3 suggests that the estimation error is sensitive to the second-order curvature of the activation manifold.

\subsubsection{Impact of Activation Smoothness}

We compare the performance of ReLU and GELU neurons\cite{gelu}. The mathematical definitions of these functions are given below:ReLU (Rectified Linear Unit):\begin{equation}\text{ReLU}(x) = \max(0, x) = \begin{cases} x, \text{if } x > 0 \\ 0, \text{if } x \leq 0 \end{cases}\end{equation}GELU (Gaussian Error Linear Unit):\begin{equation}\text{GELU}(x) = x \Phi(x) = x \cdot \frac{1}{2} \left[ 1 + \text{erf}\left( \frac{x}{\sqrt{2}} \right) \right]\end{equation}where $\Phi(x)$ is the cumulative distribution function of the standard normal distribution.

As illustrated in , the first-order derivative of ReLU is discontinuous at $x=0$, leading to a singular second-order derivative (Dirac delta distribution). In contrast, GELU provides a smooth, infinitely differentiable curve.

Our experiments confirm that GELU consistently yields higher cosine similarity than ReLU. This is because the finite and bounded second-order curvature of GELU reduces the variance of the finite-difference approximation during the hierarchical bisection, allowing HZO to capture subtle directional signals that are otherwise lost in the sharp "dead zones" of ReLU.

\subsubsection{Sensitivity to Numerical Precision}

\begin{figure}[H]
    \centering
    \begin{subfigure}[b]{0.45\textwidth}
        \centering
        \includegraphics[width=\textwidth]{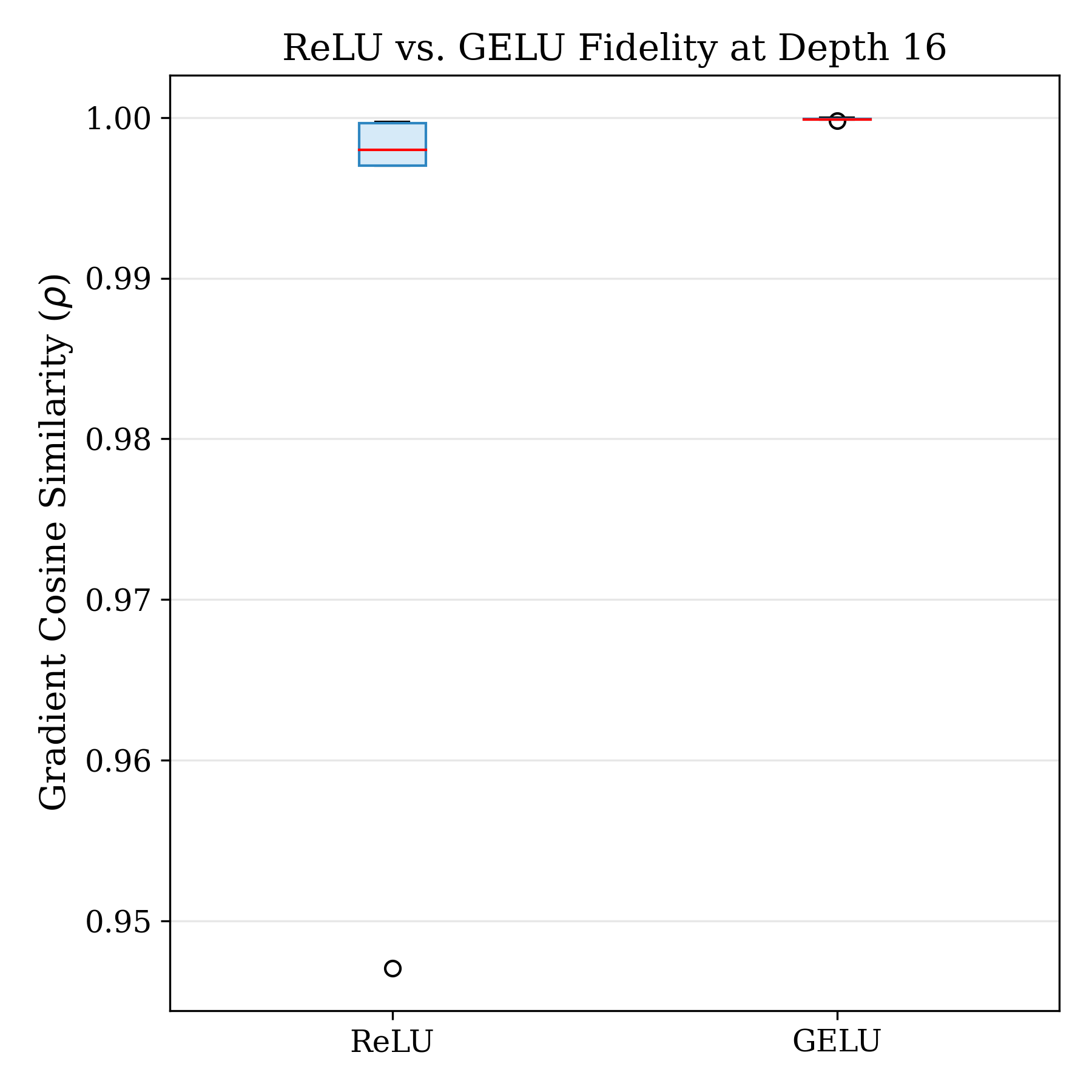}
        \caption{ReLU cs GELU}
        \label{fig:sub1}
    \end{subfigure}
    \hfill
    \begin{subfigure}[b]{0.45\textwidth}
        \centering
        \includegraphics[width=\textwidth]{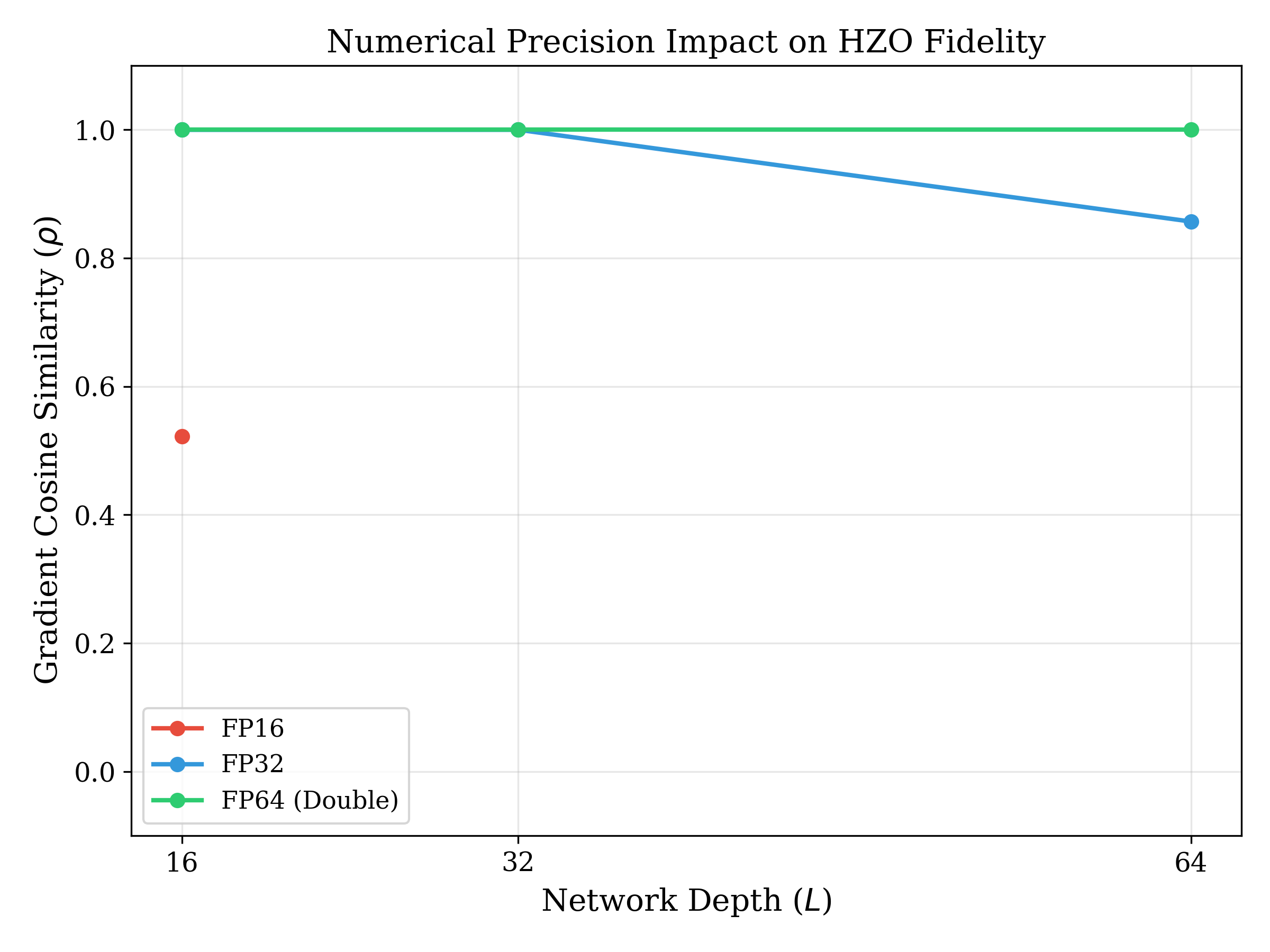}
        \caption{Numerical Precision}
        \label{fig:sub2}
    \end{subfigure}
    \caption{Experimental Results with different activation functions and numerical precision}
    \label{fig:two_images2}
\end{figure}

Numerical precision is a critical factor in zeroth-order optimization due to the subtraction of two close activation values. We evaluated HZO across different network depths using float16, float32, and float64. Results \ref{fig:two_images2} show that float32 is the minimum requirement for deep networks.

\section{Conclusion}
In this work, we have presented Hierarchical Zeroth-Order Optimization (HZO), a novel paradigm that fundamentally reimagines the mechanism of gradient estimation in deep neural networks without backpropagation. By building upon and extending the neuron-wise perturbation framework proposed by R et al. (DeepZero), we have addressed the long-standing challenges of computational inefficiency and gradient decoherence in deep zeroth-order learning.

Our core contribution, the Divide-and-Conquer optimization strategy, successfully reduces the time complexity of single gradient queries by orders of magnitude. This makes zeroth-order optimization not only a theoretical curiosity but a computationally viable alternative to first-order methods.Through theoretical analysis, we have identified the necessary conditions for the convergence of deep zeroth-order optimization. We demonstrated that Residual Connections are essential for anchoring the Lipschitz constant near unity ($L_{lip} \approx 1$), thereby preventing exponential error inflation across layers. 

Experimental results on CIFAR-10 and a 10-class subset of ImageNet validate the scalability and robustness of HZO. Achieving a competitive 74.2\% accuracy on CIFAR-10 and demonstrating stable convergence on ImageNet-10 proves that our method possesses the potential for application in large-scale model training. By bypassing the need for a global backward pass, HZO opens new avenues for exploring biologically-plausible learning and the optimization of non-differentiable hardware systems.

\bibliography{reference}

\begin{thebibliography}{10}

\bibitem{bengio2014autoencodersprovidecreditassignment}
Yoshua Bengio.
\newblock How auto-encoders could provide credit assignment in deep networks via target propagation, 2014.

\bibitem{beny2013deep}
C{\'e}dric B{\'e}ny.
\newblock Deep learning and the renormalization group.
\newblock {\em arXiv preprint arXiv:1301.3124}, 2013.

\bibitem{cao2024acceleratedgradientmethodconvex}
Jincheng Cao, Ruichen Jiang, Erfan~Yazdandoost Hamedani, and Aryan Mokhtari.
\newblock An accelerated gradient method for convex smooth simple bilevel optimization, 2024.

\bibitem{chen2023deepzero}
Aochuan Chen, Yimeng Zhang, Jinghan Jia, James Diffenderfer, Jiancheng Liu, Konstantinos Parasyris, Yihua Zhang, Zheng Zhang, Bhavya Kailkhura, and Sijia Liu.
\newblock Deepzero: Scaling up zeroth-order optimization for deep model training.
\newblock {\em arXiv preprint arXiv:2310.02025}, 2023.

\bibitem{deng2009imagenet}
Jia Deng, Wei Dong, Richard Socher, Li-Jia Li, Kai Li, and Li~Fei-Fei.
\newblock Imagenet: A large-scale hierarchical image database.
\newblock In {\em 2009 IEEE conference on computer vision and pattern recognition}, pages 248--255. Ieee, 2009.

\bibitem{frank2023hierarchicalinbiology}
Sebastian~M Frank, Marvin~R Maechler, Sergey~V Fogelson, and Peter~U Tse.
\newblock Hierarchical categorization learning is associated with representational changes in the dorsal striatum and posterior frontal and parietal cortex.
\newblock {\em Human Brain Mapping}, 44(9):3897--3912, 2023.

\bibitem{guang2}
Tingzhao Fu, Jianfa Zhang, Run Sun, Yuyao Huang, Wei Xu, Sigang Yang, Zhihong Zhu, and Hongwei Chen.
\newblock Optical neural networks: progress and challenges.
\newblock {\em Light: Science \& Applications}, 13(1):263, 2024.

\bibitem{zoguang1}
Jiaqi Gu, Chenghao Feng, Zheng Zhao, Zhoufeng Ying, Ray~T Chen, and David~Z Pan.
\newblock Efficient on-chip learning for optical neural networks through power-aware sparse zeroth-order optimization.
\newblock In {\em Proceedings of the AAAI conference on artificial intelligence}, volume~35, pages 7583--7591, 2021.

\bibitem{guang1}
Xianxin Guo, Thomas~D Barrett, Zhiming~M Wang, and AI~Lvovsky.
\newblock Backpropagation through nonlinear units for the all-optical training of neural networks.
\newblock {\em Photonics Research}, 9(3):B71--B80, 2021.

\bibitem{hanson1990stochastic}
Stephen~Jos{\'e} Hanson.
\newblock A stochastic version of the delta rule.
\newblock {\em Physica D: Nonlinear Phenomena}, 42(1-3):265--272, 1990.

\bibitem{gelu}
D~Hendrycks.
\newblock Gaussian error linear units (gelus).
\newblock {\em arXiv preprint arXiv:1606.08415}, 2016.

\bibitem{hhmodel}
Alan~L Hodgkin and Andrew~F Huxley.
\newblock A quantitative description of membrane current and its application to conduction and excitation in nerve.
\newblock {\em The Journal of physiology}, 117(4):500, 1952.

\bibitem{koslik2024hiddensemimarkovmodelsinhomogeneous}
Jan-Ole Koslik.
\newblock Hidden semi-markov models with inhomogeneous state dwell-time distributions, 2024.

\bibitem{cifar10}
Alex Krizhevsky, Geoffrey Hinton, et~al.
\newblock Learning multiple layers of features from tiny images.
\newblock 2009.

\bibitem{lillicrap2014randomfeedbackweightssupport}
Timothy~P. Lillicrap, Daniel Cownden, Douglas~B. Tweed, and Colin~J. Akerman.
\newblock Random feedback weights support learning in deep neural networks, 2014.

\bibitem{bpnotreasonable1}
Timothy~P Lillicrap, Adam Santoro, Luke Marris, Colin~J Akerman, and Geoffrey Hinton.
\newblock Backpropagation and the brain.
\newblock {\em Nature Reviews Neuroscience}, 21(6):335--346, 2020.

\bibitem{liu2022electrophysiological}
Meng Liu, Wenshan Dong, Shaozheng Qin, Tom Verguts, and Qi~Chen.
\newblock Electrophysiological signatures of hierarchical learning.
\newblock {\em Cerebral Cortex}, 32(3):626--639, 2022.

\bibitem{malladi2024finetuninglanguagemodelsjust}
Sadhika Malladi, Tianyu Gao, Eshaan Nichani, Alex Damian, Jason~D. Lee, Danqi Chen, and Sanjeev Arora.
\newblock Fine-tuning language models with just forward passes, 2024.

\bibitem{mehta2014exact}
Pankaj Mehta and David~J Schwab.
\newblock An exact mapping between the variational renormalization group and deep learning.
\newblock {\em arXiv preprint arXiv:1410.3831}, 2014.

\bibitem{mi2025fastllmfinetuningzerothorder}
Zhendong Mi, Qitao Tan, Grace~Li Zhang, Zhaozhuo Xu, Geng Yuan, and Shaoyi Huang.
\newblock Towards fast llm fine-tuning through zeroth-order optimization with projected gradient-aligned perturbations, 2025.

\bibitem{ren2022scaling}
Mengye Ren, Simon Kornblith, Renjie Liao, and Geoffrey Hinton.
\newblock Scaling forward gradient with local losses.
\newblock {\em arXiv preprint arXiv:2210.03310}, 2022.

\bibitem{1986hintonbp}
David~E Rumelhart, Geoffrey~E Hinton, and Ronald~J Williams.
\newblock Learning representations by back-propagating errors.
\newblock {\em nature}, 323(6088):533--536, 1986.

\bibitem{zoguang2}
Hiroshi Sawada, Kazuo Aoyama, and Masaya Notomi.
\newblock Layered-parameter perturbation for zeroth-order optimization of optical neural networks.
\newblock In {\em Proceedings of the AAAI Conference on Artificial Intelligence}, volume~39, pages 20292--20301, 2025.

\bibitem{bpnotreasonable2}
Navid Shervani-Tabar and Robert Rosenbaum.
\newblock Meta-learning biologically plausible plasticity rules with random feedback pathways.
\newblock {\em Nature Communications}, 14(1):1805, 2023.

\bibitem{guang3}
James Spall, Xianxin Guo, and Alexander~I Lvovsky.
\newblock Training neural networks with end-to-end optical backpropagation.
\newblock {\em Advanced Photonics}, 7(1):016004--016004, 2025.

\bibitem{theves2021learning}
Stephanie Theves, David~A Neville, Guill{\'e}n Fern{\'a}ndez, and Christian~F Doeller.
\newblock Learning and representation of hierarchical concepts in hippocampus and prefrontal cortex.
\newblock {\em Journal of Neuroscience}, 41(36):7675--7686, 2021.

\bibitem{yu2024subzero}
Ziming Yu, Pan Zhou, Sike Wang, Jia Li, and Hua Huang.
\newblock Subzero: Random subspace zeroth-order optimization for memory-efficient llm fine-tuning.
\newblock 2024.

\bibitem{2024llmfinetuning}
Yihua Zhang, Pingzhi Li, Junyuan Hong, Jiaxiang Li, Yimeng Zhang, Wenqing Zheng, Pin-Yu Chen, Jason~D Lee, Wotao Yin, Mingyi Hong, et~al.
\newblock Revisiting zeroth-order optimization for memory-efficient llm fine-tuning: A benchmark.
\newblock {\em arXiv preprint arXiv:2402.11592}, 2024.

\bibitem{zolianghua}
Jiajun Zhou, Yifan Yang, Kai Zhen, Ziyue Liu, Yequan Zhao, Ershad Banijamali, Athanasios Mouchtaris, Ngai Wong, and Zheng Zhang.
\newblock Quzo: Quantized zeroth-order fine-tuning for large language models.
\newblock {\em arXiv preprint arXiv:2502.12346}, 2025.

\end{thebibliography}
\bibliographystyle{plain}

\newpage
\appendix
\section{Proof of Theorem 3}
\label{appendix:proof_theorem3}

In this section, we provide a formal derivation of the gradient estimation error for the Hierarchical Zeroth-Order (HZO) optimizer. We show that the estimation error of HZO matches the \emph{worst-case scaling behavior} of direct end-to-end zeroth-order (ZO) methods with respect to network depth and smoothness.

\subsection{Preliminaries and Single-step JVP Error}

Consider an $L$-layer neural network
$
F(\mathbf{x}) = f_L \circ f_{L-1} \circ \dots \circ f_1(\mathbf{x}),
$
where $\mathbf{a}_l$ denotes the activation at layer $l$, with $\mathbf{a}_0 = \mathbf{x}$. Let $\mathcal{L}$ be the loss function. The true gradient satisfies the chain rule
$
\mathbf{g}_{l-1} = \mathbf{J}_l^T \mathbf{g}_l,
$
where $\mathbf{J}_l = \partial f_l / \partial \mathbf{a}_{l-1}$.

HZO estimates vector--Jacobian products of the form $\mathbf{J}_l^T \mathbf{v}$ using symmetric finite differences. For a perturbation magnitude $\delta > 0$ and a direction vector $\mathbf{u}$, the estimator is given by
\begin{equation}
\hat{\mathbf{J}}_l^T \mathbf{v}
=
\mathbb{E}_{\mathbf{u}}
\left[
\frac{f_l(\mathbf{a}_{l-1} + \delta \mathbf{u}) - f_l(\mathbf{a}_{l-1} - \delta \mathbf{u})}{2\delta}
(\mathbf{u}^\top \mathbf{v})
\right],
\end{equation}
where the expectation is taken over either random directions or a coordinate basis.

Assuming $f_l \in \mathcal{C}^3$ with Hessian Lipschitz constant $\beta_l$, a Taylor expansion around $\mathbf{a}_{l-1}$ yields
\begin{equation}
\hat{\mathbf{J}}_l^T \mathbf{v}
=
\mathbf{J}_l^T \mathbf{v}
+
\delta^2 \mathcal{E}_l(\mathbf{v})
+
\mathcal{O}(\delta^4),
\end{equation}
where $\|\mathcal{E}_l(\mathbf{v})\| \le C \beta_l \|\mathbf{v}\|$ for a constant $C$ depending on third-order derivatives. Hence, the local estimation error satisfies
\begin{equation}
\|\hat{\mathbf{J}}_l^T \mathbf{v} - \mathbf{J}_l^T \mathbf{v}\|
=
\mathcal{O}(\beta_l \delta^2).
\end{equation}

\subsection{Recursive Error Propagation}

Let $\mathbf{g}_l$ denote the true gradient at layer $l$ and $\hat{\mathbf{g}}_l$ the corresponding HZO estimate. Define the estimation error $\mathbf{e}_l = \hat{\mathbf{g}}_l - \mathbf{g}_l$. The recursive update of HZO can be written as
\begin{equation}
\hat{\mathbf{g}}_{l-1}
=
(\mathbf{J}_l^T + \Delta \mathbf{J}_l^T)(\mathbf{g}_l + \mathbf{e}_l),
\end{equation}
where $\Delta \mathbf{J}_l^T$ denotes the finite-difference Jacobian estimation error.

Subtracting the true recursion $\mathbf{g}_{l-1} = \mathbf{J}_l^T \mathbf{g}_l$ yields
\begin{equation}
\mathbf{e}_{l-1}
=
\mathbf{J}_l^T \mathbf{e}_l
+
\Delta \mathbf{J}_l^T \mathbf{g}_l
+
\Delta \mathbf{J}_l^T \mathbf{e}_l .
\end{equation}
Taking norms and applying triangle inequality gives
\begin{equation}
\|\mathbf{e}_{l-1}\|
\le
\|\mathbf{J}_l^T\| \, \|\mathbf{e}_l\|
+
\|\Delta \mathbf{J}_l^T \mathbf{g}_l\|
+
\|\Delta \mathbf{J}_l^T\| \, \|\mathbf{e}_l\|.
\end{equation}
Using $\|\mathbf{J}_l^T\| \le L_{\mathrm{lip}}$ and $\|\Delta \mathbf{J}_l^T\| \le C \beta \delta^2$, we obtain
\begin{equation}
\|\mathbf{e}_{l-1}\|
\le
(L_{\mathrm{lip}} + C \beta \delta^2)\|\mathbf{e}_l\|
+
C \beta \delta^2 \|\mathbf{g}_l\|.
\end{equation}

\subsection{Total Cumulative Error}

Iterating the above bound from $l=L$ to $l=0$ yields
\begin{equation}
\|\mathbf{e}_0\|
\le
C \beta \delta^2
\sum_{l=1}^{L}
(L_{\mathrm{lip}} + C \beta \delta^2)^{l-1}
\|\mathbf{g}_l\|.
\end{equation}
Assuming a standard worst-case bound on gradient propagation,
\begin{equation}
\|\mathbf{g}_l\|
\le
C_g \|\mathbf{g}_L\| L_{\mathrm{lip}}^{L-l},
\end{equation}
we obtain
\begin{equation}
\|\mathbf{e}_{\mathrm{HZO}}\|
=
\mathcal{O}
\left(
\beta \delta^2
\frac{L_{\mathrm{lip}}^L - 1}{L_{\mathrm{lip}} - 1}
\right),
\end{equation}
which matches the worst-case depth and smoothness dependence observed in direct end-to-end zeroth-order gradient estimation.

\subsection{Stability in the Unitary Limit}

The above result highlights the importance of the \textbf{unitary limit} $L_{\mathrm{lip}} \approx 1$:
\begin{itemize}
\item \textbf{Case $L_{\mathrm{lip}} > 1$:} The estimation error grows exponentially with depth, rendering gradient signals unstable.
\item \textbf{Case $L_{\mathrm{lip}} \approx 1$:} Using
\(
\lim_{L_{\mathrm{lip}} \to 1}
\frac{L_{\mathrm{lip}}^L - 1}{L_{\mathrm{lip}} - 1}
= L,
\)
the error scales linearly with depth:
\begin{equation}
\|\mathbf{e}_{\mathrm{HZO}}\|
=
\mathcal{O}(L \beta \delta^2).
\end{equation}
\end{itemize}
This explains why architectures with residual connections, which naturally operate near the unitary regime at initialization, exhibit stable and high-fidelity gradient estimates under HZO even for deep networks.

\end{document}